\documentclass[10pt,twocolumn,letterpaper]{article}

\usepackage{cvpr}
\usepackage{times}
\usepackage{epsfig}
\usepackage{graphicx}
\usepackage{amsmath}
\usepackage{amssymb}
\usepackage{float}
\usepackage{subcaption}
\usepackage{afterpage}

\usepackage[breaklinks=true,bookmarks=false]{hyperref}

\cvprfinalcopy % *** Uncomment this line for the final submission

\def\cvprPaperID{****} % *** Enter the CVPR Paper ID here

\begin{document}

%%%%%%%%% TITLE
\title{Rethinking Radiology: An Analysis of Different Approaches to BraTS}

\author{William Bakst\thanks{Equal Contribution}\\
Stanford University\\
{\tt\small wbakst@stanford.edu}
\and
Linus Meyer-Teruel\footnotemark[1]\\
Stanford University\\
{\tt\small lmeyerte@stanford.edu}
\and
Jasdeep Singh\footnotemark[1]\\
Stanford University\\
{\tt\small jasdeep@stanford.edu}
}

\maketitle
%\thispagestyle{empty}

%%%%%%%%% ABSTRACT
\begin{abstract}
   This paper discusses the deep learning architectures currently used for pixel-wise segmentation of primary and secondary glioblastomas and low-grade gliomas.  We implement various models such as the popular UNet architecture and compare the performance of these implementations on the BRATS dataset. This paper will explore the different approaches and combinations, offering an in depth discussion of how they perform and how we may improve upon them using more recent advancements in deep learning architectures. 
\end{abstract}

%%%%%%%%% Introduction
\section{Introduction} \label{introduction}
Cancer is one of the leading causes of death in the world today, responsible for hundreds of thousands of deaths each year.  With the rise of deep learning architectures, particularly Convolutional Neural Networks \cite{all_conv}, computers have become incredibly powerful resources for the improvement of our medical abilities. One very important improvement is in detecting diseases early on, which is one of the most effective preventative measure that we currently have for cancer.  

The specific problem that we approach is the pixel-wise segmentation of primary and secondary glioblastomas and low-grade gliomas (see section \ref{data} for more information on the dataset to be used).  Systems for such segmentation enable doctors to waste less time diagnosing medical images and more time saving their patients.  There has been much research focused on improving our computer vision systems \cite{BraTS2,unet3d,senet,BraTS,inception,neXt,ures}, which we can use for the early detection of glioblastoma from MRI scans.  While current results are quite good, there is still large room for improvement.

Our approach to this problem has two steps.  First, we implement some of the current state-of-the-art models used for this task (see above citation).  We then train and test these models, performing experiments and evaluating the results for each model.  These models take advantage of recent advancements in deep learning and perform rather well.  Second, given the many different approaches and architectures, we believe that it is worth taking the time to test each architecture and perform in-depth analysis and comparisons of the models.  In doing so, we believe that we can help improve these systems by better understanding their results and learning which approaches are the most effective, ultimately enabling further research on this topic.  Thus, this paper focuses on the implementation and training of these different architectures while providing an analysis of the results of our experiments for each one as well as comparisons between each of the models (see section \ref{eval} for more details on evaluation metrics).

%%%%%%%%% Challenges
\section{Challenges} \label{challenges}
There are a few challenges regarding our problem that we must discuss.  First is that MRI data is three dimensional and we must use architectures that can process it in that dimension space. Furthermore, because MRI scans take approximately 40 minutes and patients tend to move around during the scan, image quality is prone to blurriness and degeneration.  While the architectures alone address this issue, we believe that preprocessing will be very useful in remedying these problems as well. However, image preprocessing can also pose a challenge. Some images can be as large as 1GB, and the preprocessing of such an image is computationally expensive. Thus, we must bound our images such that our computers can still handle the images without limiting the ability of our models.

%%%%%%%%% Relevant Research
\section{Related Work} \label{related}

With the introduction of AlexNet \cite{alexnet} in 2012, Convolutional Neural Networks (CNN's) became the defacto architectures for visual tasks. Since then we have seen an explosion of different architectures and techniques that built upon the vanilla convolutional design. First we saw benefits from increasing depth in the VGGNets \cite{vgg} and Inception modules \cite{inception}. Then ResNet \cite{resnet} displayed how to build even deeper and more effective CNNs through the use of identity based skip connections. Highway networks \cite{highway} use gated skip connections between layers to help training and to build deeper networks. Both residual connections and highway connections are specific instances of more general interlayer connections that have been shown to improve the expressive ability and performance of CNNs. Not only do these connections improve performance but they also greatly increase the stability and ease of training deep CNN's \cite{inception}. 

Theoretically, residual connections can be thought of as identity branches of a multi-branch convolutional module. The Inception modules are also examples of successful multi-branch architectures. Recently, the work behind the ResNeXt architecture \cite{neXt} shows gains of using multi-branch architectures for object detection and instance segmentation. Even more recently, Hu et al. \cite{senet} show that there Squeeze-and-Excitation (SE) module can be used with all of these advances in architecture design to give performance boosts in object detection. The SE module works by re-weighting filter channels conditioned on the global distribution of channels for that CNN layer. This seems to give performance boosts for object detection; however, to our knowledge, it has yet to be tested on segmentation tasks. 
    
This seems to be a general trend in architecture selection. Many advances in CNN architectures have often only been developed on detection tasks. Even the Mask R-CNN \cite{mask_rcnn}, which has been shown to achieve state-of-the-art results for many tasks including instance segmentation, bounding box object detection, and person key-point detection, is based on an architecture which was initially developed for object detection. The Mask R-CNN simply replaces the last few layers of the network to develop new capabilities. This trend to develop and validate architecture enhancements on object detection tasks may be part due to availability of large detection datasets. However, this leaves an open question as to how many of these architectural innovations will transfer over to architectures built specifically for segmentation tasks, like the UNet \cite{unet3d}, in domains with relatively limited amounts of labeled data, such as Brain Tumor Segmentation (See section \ref{data} below). To this extent, we evaluate the effect of many of these recent architectural advancements on the UNet3D as it pertains to Brain Tumor Segmentation. 

%%%%%%%%% Datasets and Features
\section{Dataset and Features} \label{data}
The Brain Tumor Segmentation Challenge \cite{BraTS,BraTS2} is a competition aimed at pushing the improvement of deep learning architectures for Brain Tumor Segmentation. Participants build models to segment and classify acute glioblastomas, including both primary and secondary tumors.  We are currently testing our models on the dataset associated with the 2017 BraTS competition and hope to train and test them on the newly released 2018 dataset. The BRATS dataset includes multiple imaging modalities such as T1 and T2 MRI tissue contrasts, T2 FLAIR, and T1 contrast-enhanced MRI, which contain a variety of phenotypes such as primary vs. secondary tumors and solid vs. infiltrative growing tumor profiles. In total, the data consists of MRI scans from 265 patients, of which 200 are high grade gliomas and 65 are low grade gliomas. Each patient has between 2-4 images associated with their scan, often comprised of all four imaging modalities listed above.  All images are stored as NII files. For each training image there is a ground-truth pixel label with five possible classes: necrosis, edema, non-enhancing tumor, enhancing tumor, and healthy tissues. The label data was hand-labeled by professional radiologists.

\begin{figure}[H]
\begin{center}
	\begin{subfigure}{.2\textwidth}
		\includegraphics[scale=0.4]{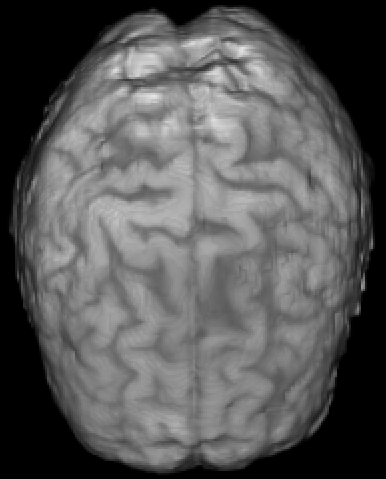}
		\label{fig:3d_top}
	\end{subfigure}
	\begin{subfigure}{.2\textwidth}
		\includegraphics[scale=0.4]{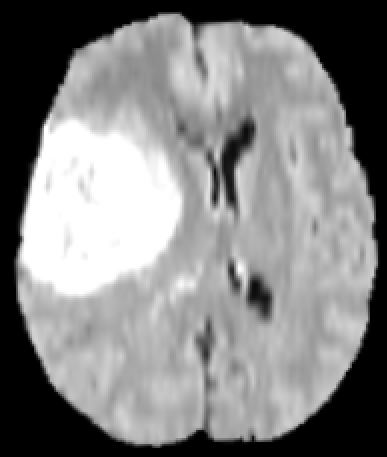}
		\label{fig:tumor_top}
	\end{subfigure}
    \begin{subfigure}{.2\textwidth}
		\includegraphics[scale=0.4]{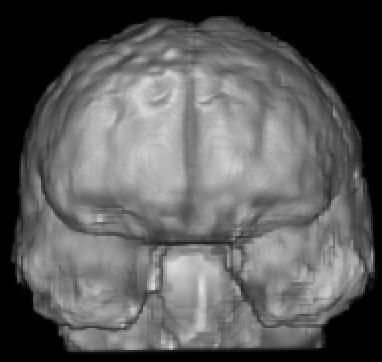}
		\label{fig:3d_front}
	\end{subfigure}
	\begin{subfigure}{.2\textwidth}
		\includegraphics[scale=0.4]{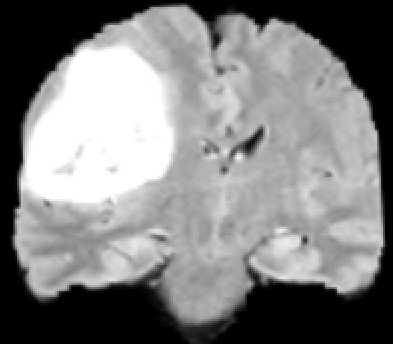}
		\label{fig:tumor_top}
	\end{subfigure}
\end{center}
\caption{MRI of a brain with a tumor from different angles}
\label{fig:mri_images}
\end{figure}

%%%%%%%%% Architectures
\section{Architectures} \label{approach}

Below we give a general overview of the architectures that we have researched and implemented. All of the architectures except the baseline are based on the original UNet3D architecture \cite{unet3d}

%-------------------------------------------------------------------------
\subsection{Evaluation} \label{eval}

We evaluate our models using a few different metrics. First, we use F1 score, which can be defined as 

\begin{align*}
\dfrac{2 \mid P \mid \mid T \mid}{\mid P \mid + \mid T \mid}
\end{align*}

where $P$ represents our prediction and $T$ represents the true label.  This scoring scheme works quite well as an estimate to the performance quality of our models because it is equally important to label a pixel as cancerous as it is to label it healthy. Second, we will be comparing the training time of our models. This is important because iterating on and updating models can take too long if training is slow. Third, we will be comparing the space complexity of each model, primarily the total number of parameters. This is also important since an incredibly large model will not necessarily be able to fit on certain hardware.

%%%%%%%%% TECHNICAL APPROACH
\subsection{Technical Approach} \label{approach}

We are researching and implementing many different methods and architectures which we describe here:

%-------------------------------------------------------------------------
\subsubsection{Baseline} \label{baseline}

Our baseline model is a simple 3 layer CNN. Each layer is CONV--MAXPOOL--ReLU where we use same padding for both the convolutional and the max pooling operations.

\begin{figure}[H]
	\centering
	\includegraphics[scale=0.32]{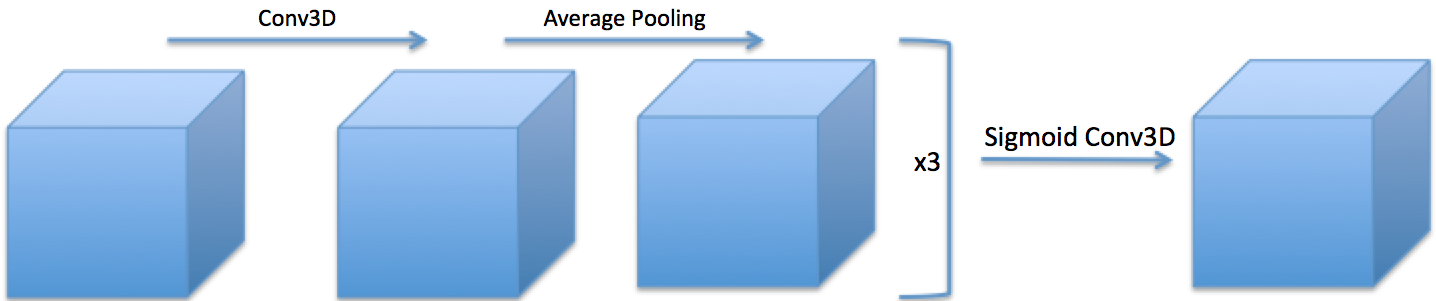}
    \caption{Baseline Model Graph}
  \label{fig:baseline}
\end{figure}

%-------------------------------------------------------------------------
\subsubsection{UNet3D} \label{unet3d}

Our UNet-3D model is built based on the original UNet3D paper \cite{unet3d}. We use the same number of standard convolutional layers, maxpools, and upconvolutions (15, 4, 4 respectively) for a total of 23 layers and 19 million parameters. Before each max-pooling, we have a double convolutional layer of the form CONV-ReLU-BATCHNORM-CONV-ReLU-BATCHNORM.

\begin{figure}[H]
	\centering
	\includegraphics[scale=0.35]{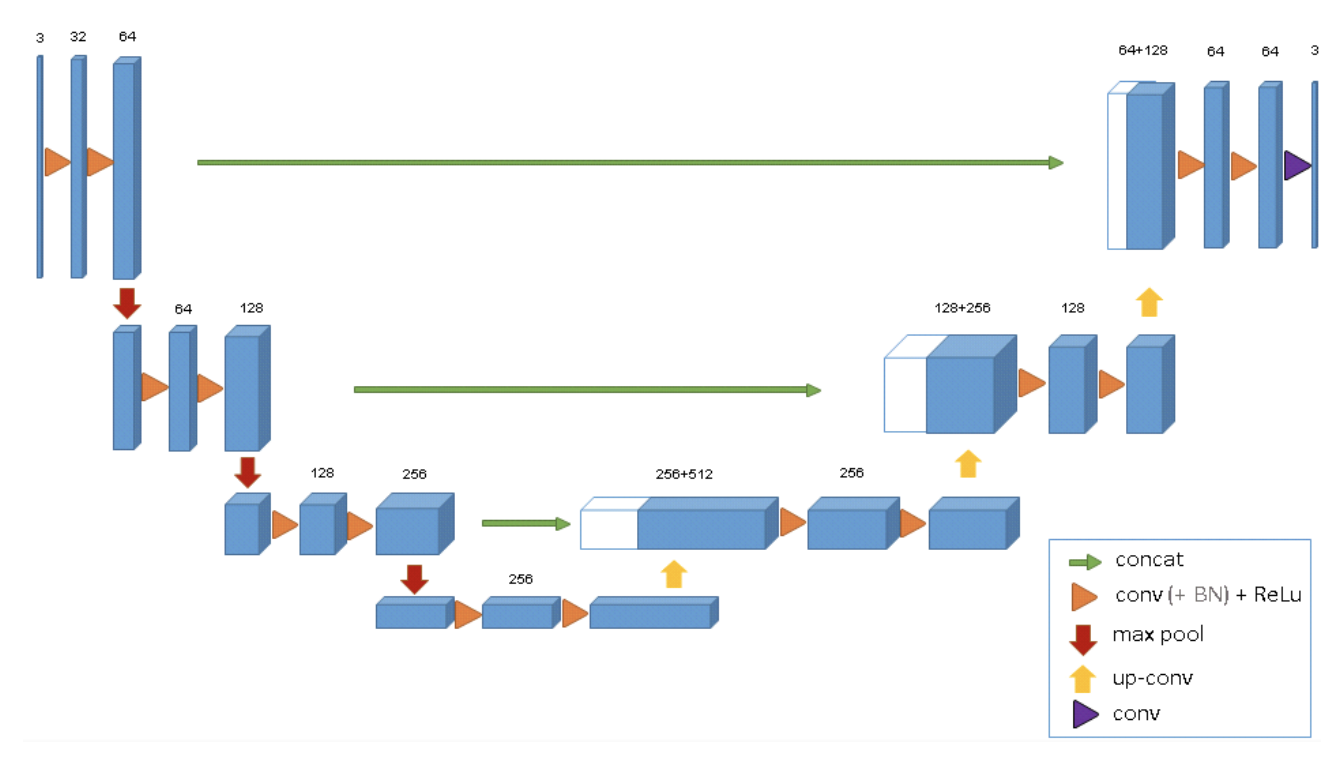}
    \caption{UNet Model Graph \cite{unet3d}}
  \label{fig:unet3d}
\end{figure}

%-------------------------------------------------------------------------
\subsubsection{UNet3D with Inception} \label{unet3d_incept}

For this model we have altered the original UNet3D architecture by replacing the double convolutional layers with single inception layers \cite{inception}. This increases the total convolutions to 42, keeps the number of maxpools and upconvolutions constant, and decreases the total number of parameters to approximately 2 million.

\begin{figure}[H]
	\centering
	\includegraphics[scale=0.35]{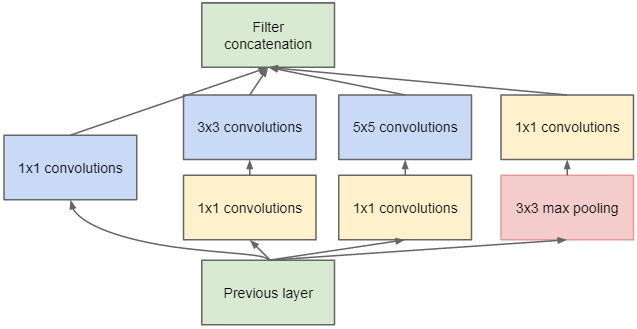}
    \caption{Inception Layer \cite{inception}}
  \label{fig:inception}
\end{figure}

%-------------------------------------------------------------------------
\subsubsection{SE-UNet3D} \label{se_unet}

For this model we have altered the original UNet3D architecture by adding squeeze-and-excitation layers for each of the convolutional blocks \cite{senet}. These connections are fully connected and add very few new parameters while enabling the model to potentially increase its complexity and learn more.

\begin{figure}[H]
	\centering
	\includegraphics[scale=0.4]{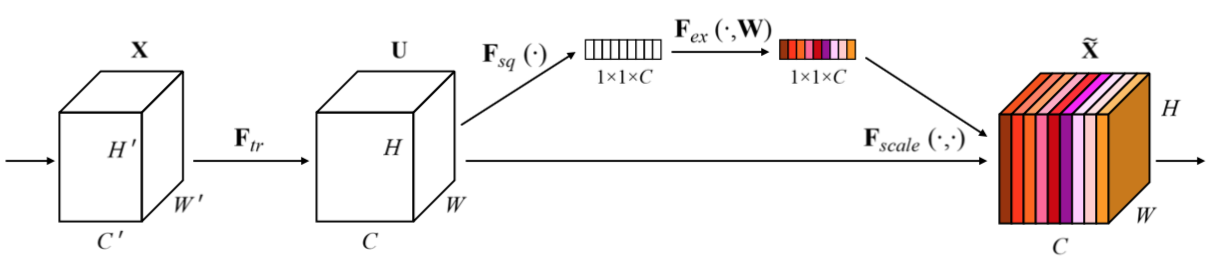}
    \caption{Squeeze-and-Excitation Module \cite{senet}}
  \label{fig:se_unet}
\end{figure}

%-------------------------------------------------------------------------
\subsubsection{SE-UNet3D with Inception} \label{se_unet_incept}

This model is the previously described UNet3D with inception layers as well as with added squeeze-and-excitation layers. The SE layers are added after the inception layers.

%-------------------------------------------------------------------------
\subsubsection{UResNet3D} \label{ures}

\begin{figure}[H]
	\centering
	\includegraphics[scale=0.6]{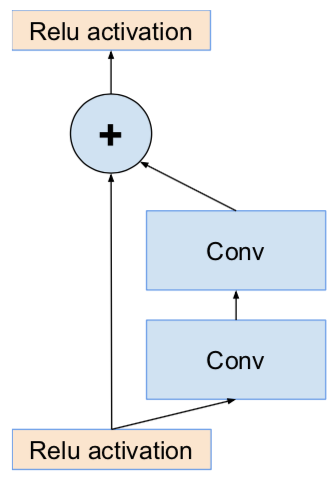}
    \caption{Example Residual Connection \cite{resnet}}
  \label{fig:residual}
\end{figure}

For this model we have altered the original UNet3D architecture by adding residual connections for each of the convolutional blocks \cite{ures}. We use convolutions with a kernel size of 1 on the input in order to increase its number of filters when adding it to the output. This increases the total number of parameters to approximately 26 million.

\begin{figure}[H]
	\centering
	\includegraphics[scale=0.6]{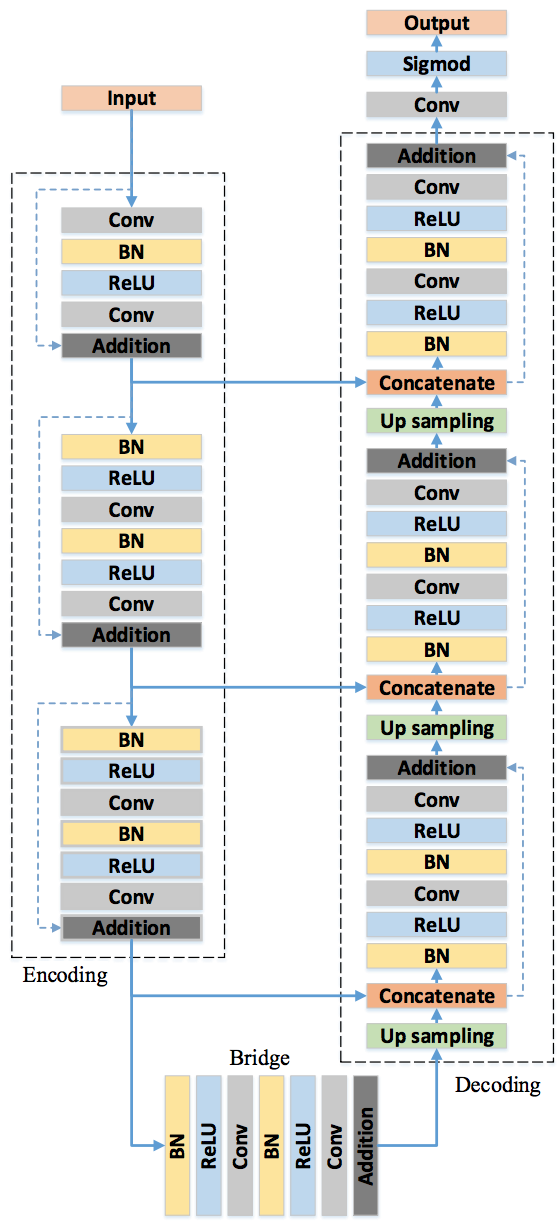}
    \caption{UNet3D with Residual Connections \cite{ures}}
  \label{fig:uresnet}
\end{figure}

%-------------------------------------------------------------------------
\subsubsection{SE-UResNet3D} \label{se_uresnet}

This model is the previously described UResNet3D with added squeeze-and-excitation layers. The SE layers are added during the convolutional blocks before the residual connection.

%-------------------------------------------------------------------------
\subsubsection{UNeXt3D} \label{unext}

For this model we have altered the original UNet3D architecture by adding aggregated transformations for each of the convolutional blocks \cite{neXt}. This layer simply repeats the same convolution for the same layer and then adds these repeated calculations together. We can think of this layer as multiple branches of the same convolutions performed on the same input. Note that the dimensions below may be different given different input dimensions for particular datasets.

\begin{figure}[H]
	\centering
	\includegraphics[scale=1.0]{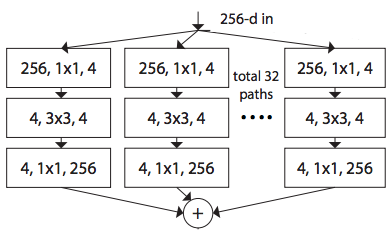}
    \caption{Example Aggregated Transformation \cite{neXt}}
  \label{fig:unext}
\end{figure}

%-------------------------------------------------------------------------
\subsubsection{UNeXt3D with Inception} \label{unext_incept}

This model is the previously described UNeXt3D with added inception layers. These inception layers are applied to each branch of the aggregated transformation.

%%%%%%%%% Experiments
\section{Experiments} \label{experiments}

Below we describe and analyze the results from our experiments for each model previously described in section \ref{approach}.  

%----------------------------------------------------------------------
\subsection{Baseline} \label{baseline_exp}

Our baseline model performed decently well, achieving a validation F1 of 0.433. This demonstrated that it was capable of finding a decent fit but was held back because of a lack in model complexity.

%----------------------------------------------------------------------
\subsection{UNet3D} \label{unet3d_exp}

The UNet3D model is based directly on the original paper, using a similar set of hyperparameters. It was one of our best performing models, which we believe is likely due to its complexity and a good set of initial hyperparameters. As we notice a degree of overfitting with a training loss significantly lower than the validation loss, we also think additions like batch normalization and dropout might improve the overall performance. 

%----------------------------------------------------------------------
\subsection{UNet3D with Inception} \label{unet3d_incept_exp}

The UNet3D model with inception layers performed slightly better than the original UNet3D in our initial tests on the 2017 dataset. It also managed to do this while using 10x fewer parameters and 30 percent less time per epoch, which demonstrates a useful improvement over the original UNet3D model.

\afterpage{%
\begin{figure*}
\begin{tabular}{ccc}
\subcaptionbox{Baseline}{\includegraphics[scale=0.23]{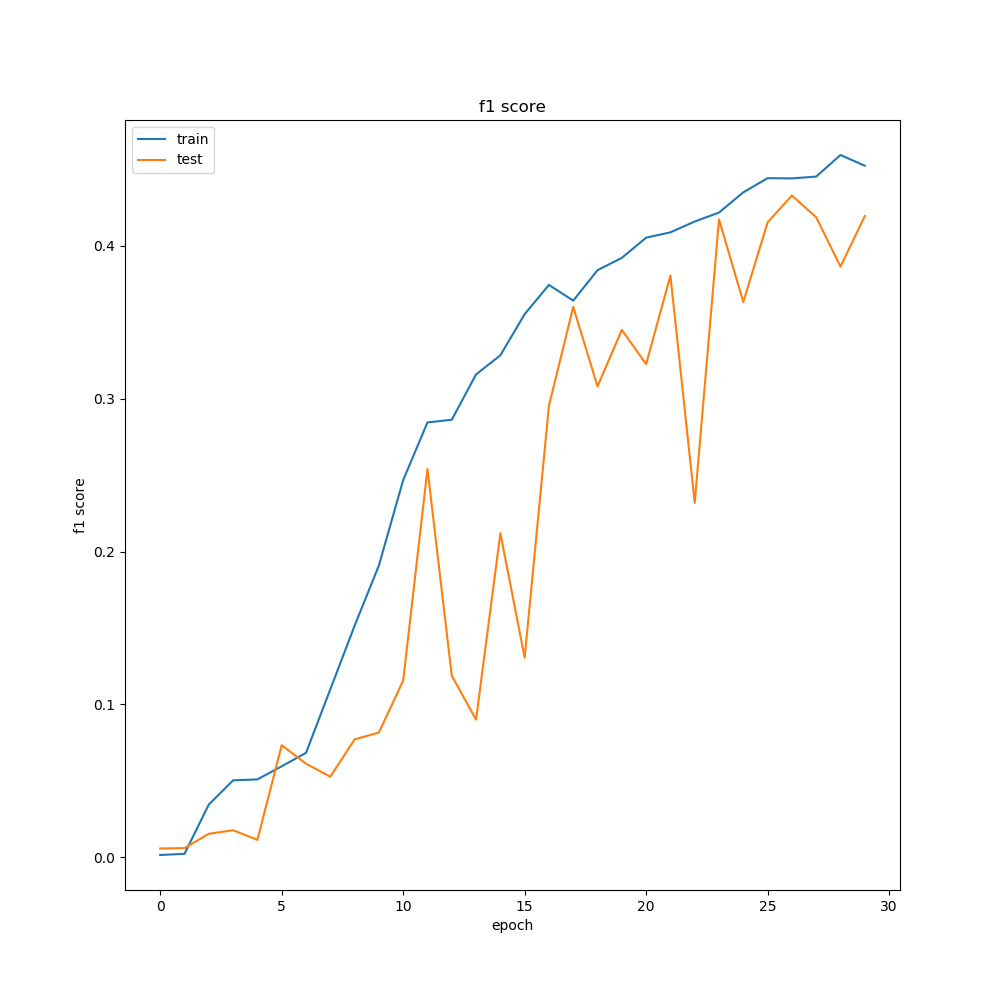}} &
\subcaptionbox{UNet3D}{\includegraphics[scale=0.23]{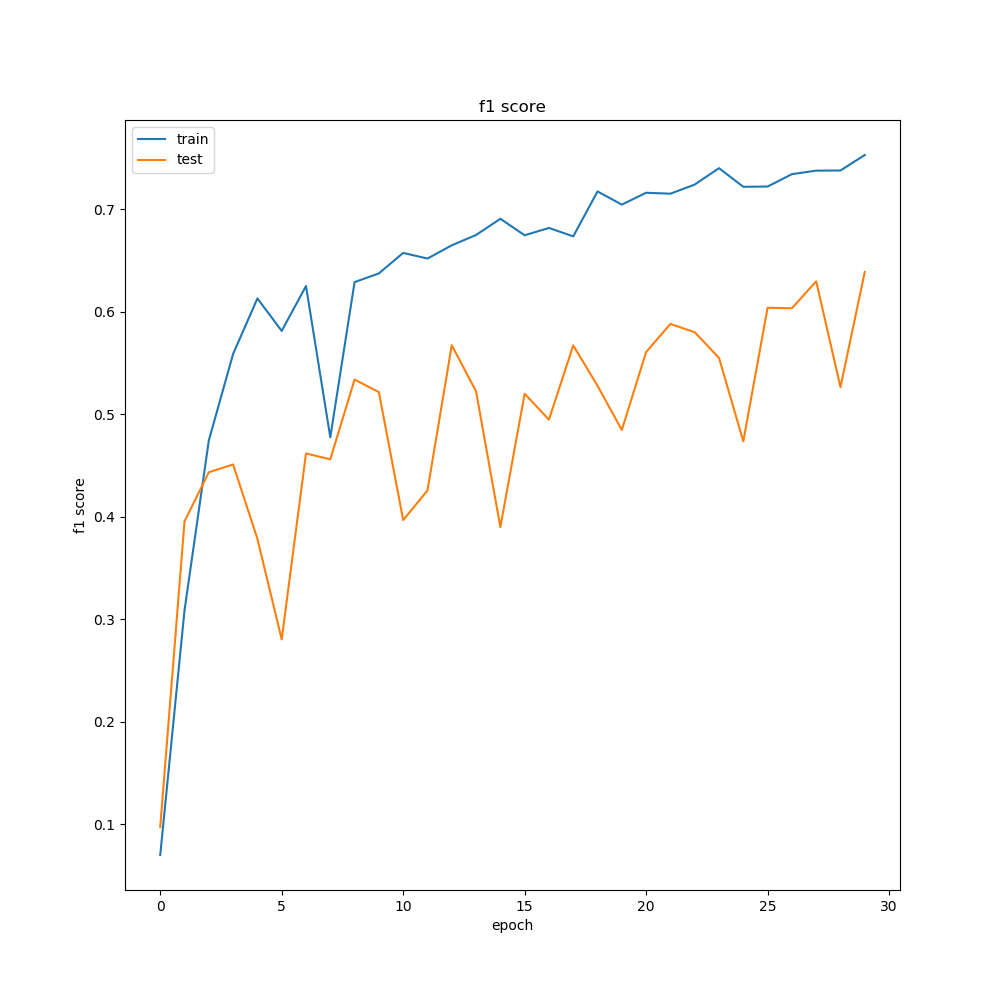}} &
\subcaptionbox{UNet3D with Inception}{\includegraphics[scale=0.23]{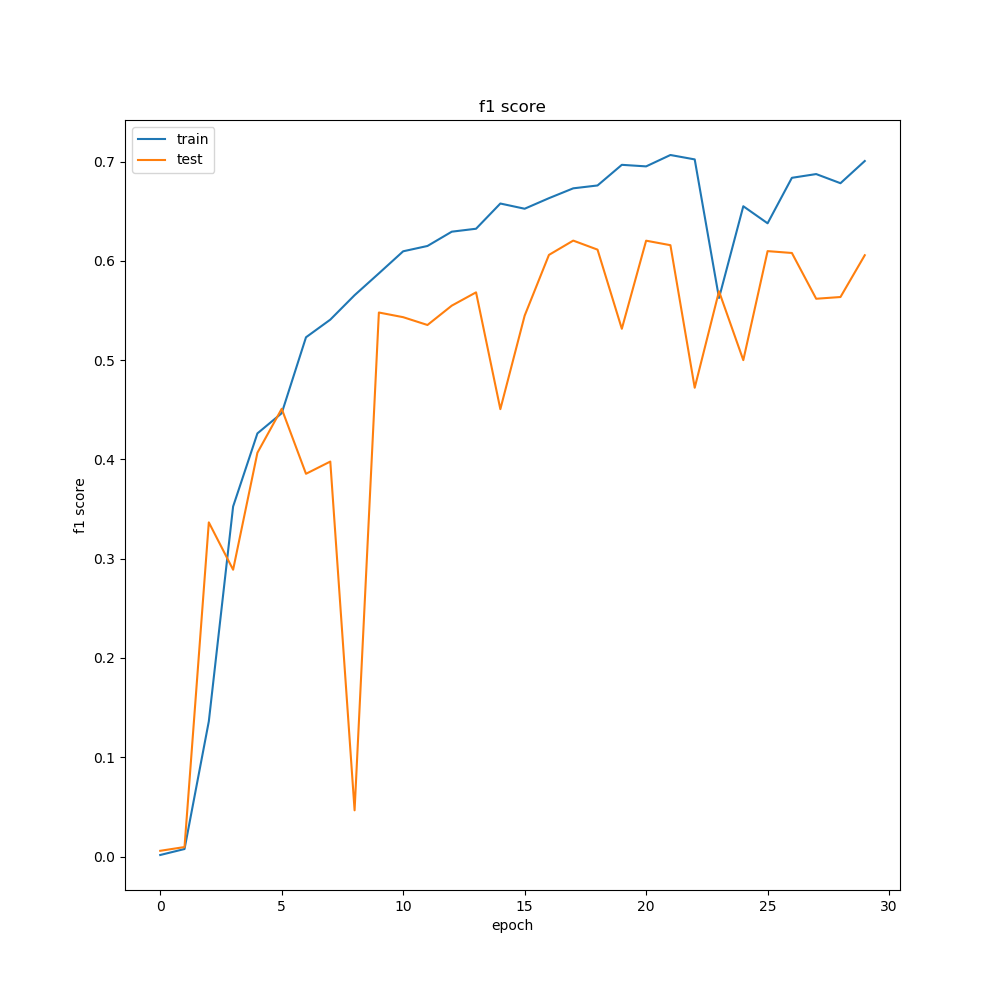}}\\
\subcaptionbox{UResNet3D}{\includegraphics[scale=0.23]{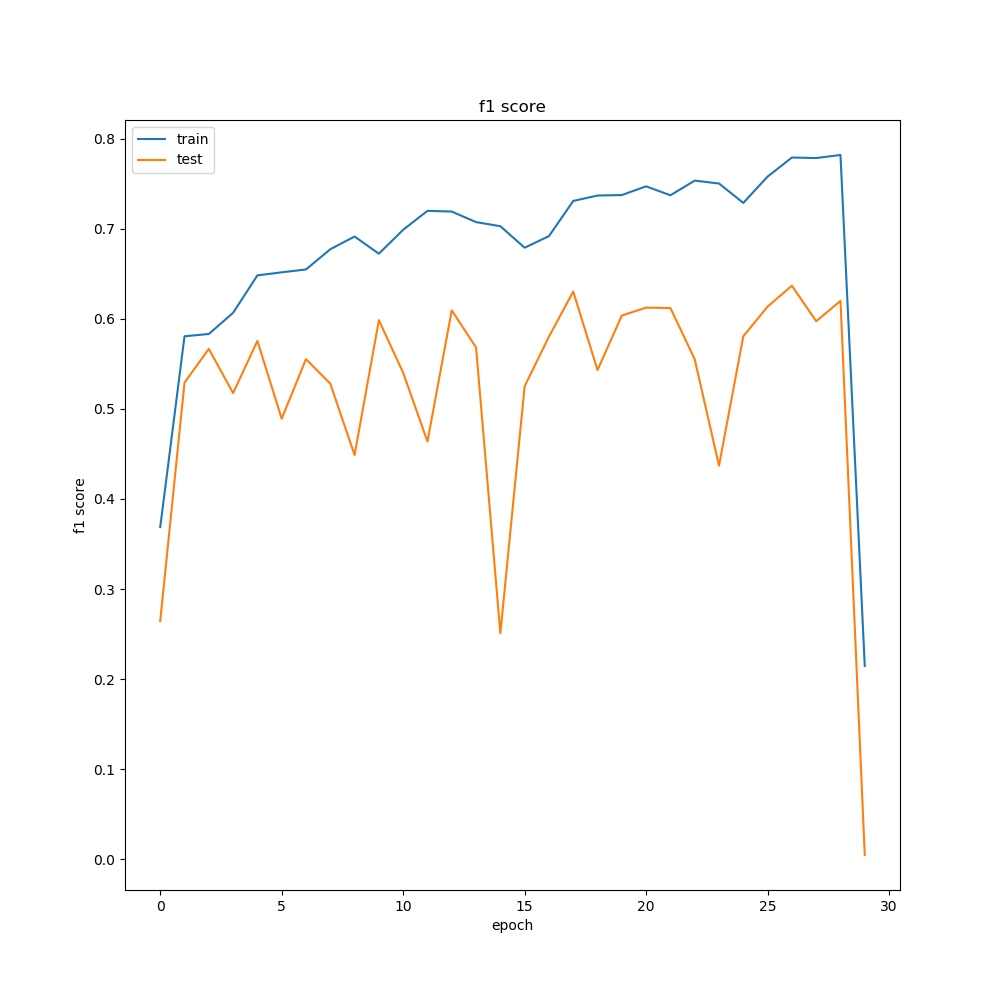}} &
\subcaptionbox{SE-UNet3D}{\includegraphics[scale=0.23]{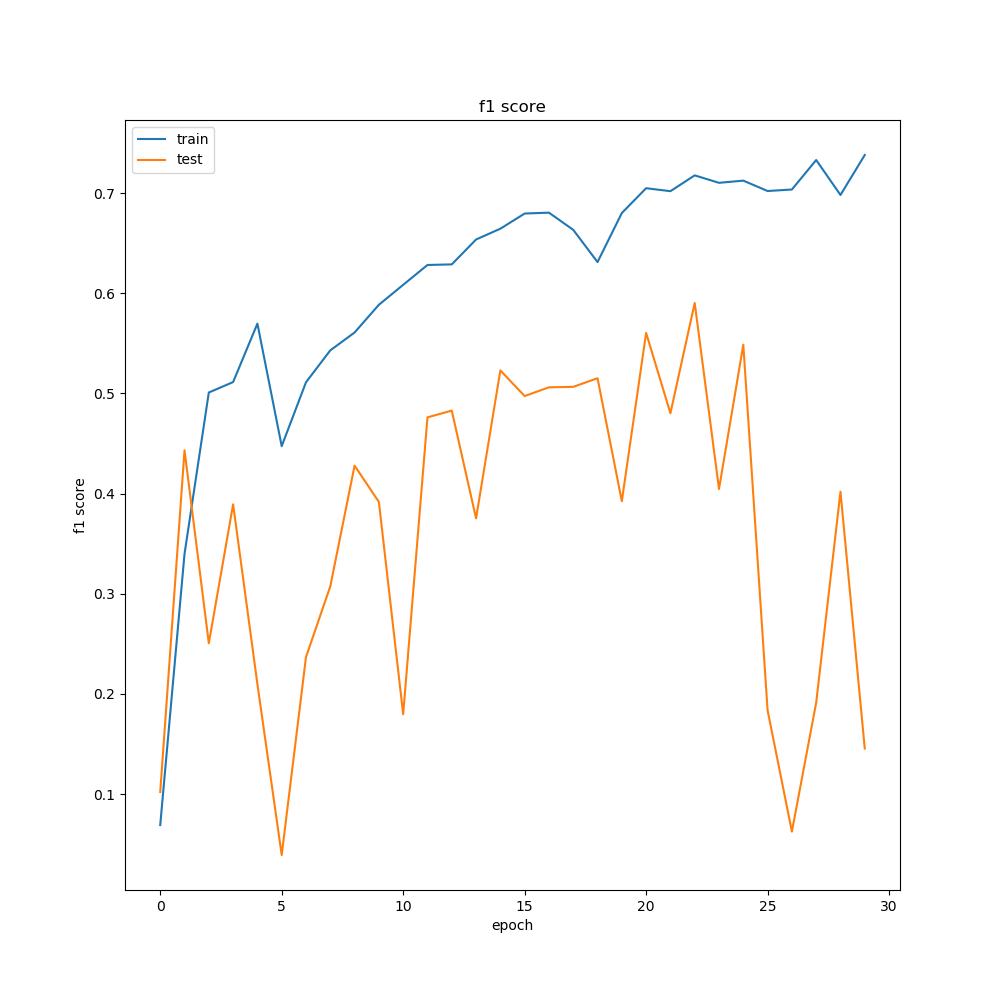}} &
\subcaptionbox{SE-UNet3D with Inception}{\includegraphics[scale=0.23]{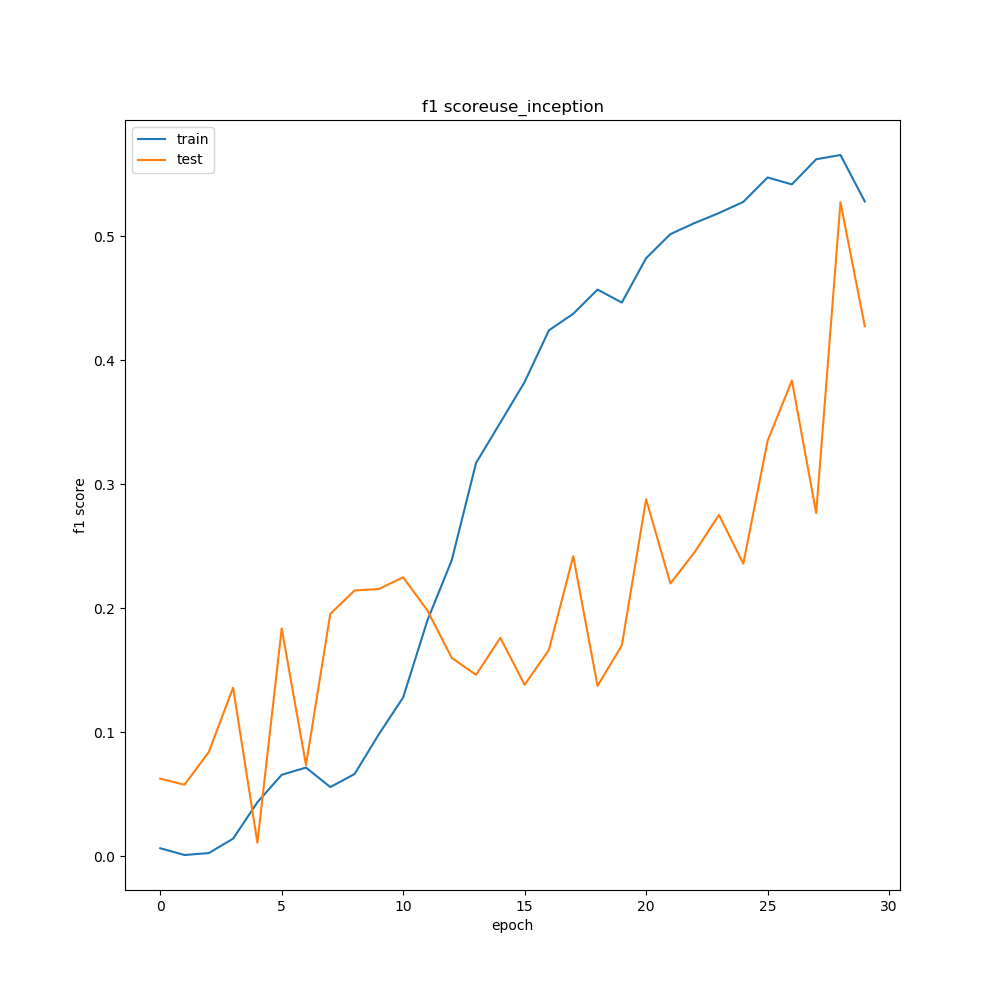}}\\
\subcaptionbox{SE-UResNet3D}{\includegraphics[scale=0.23]{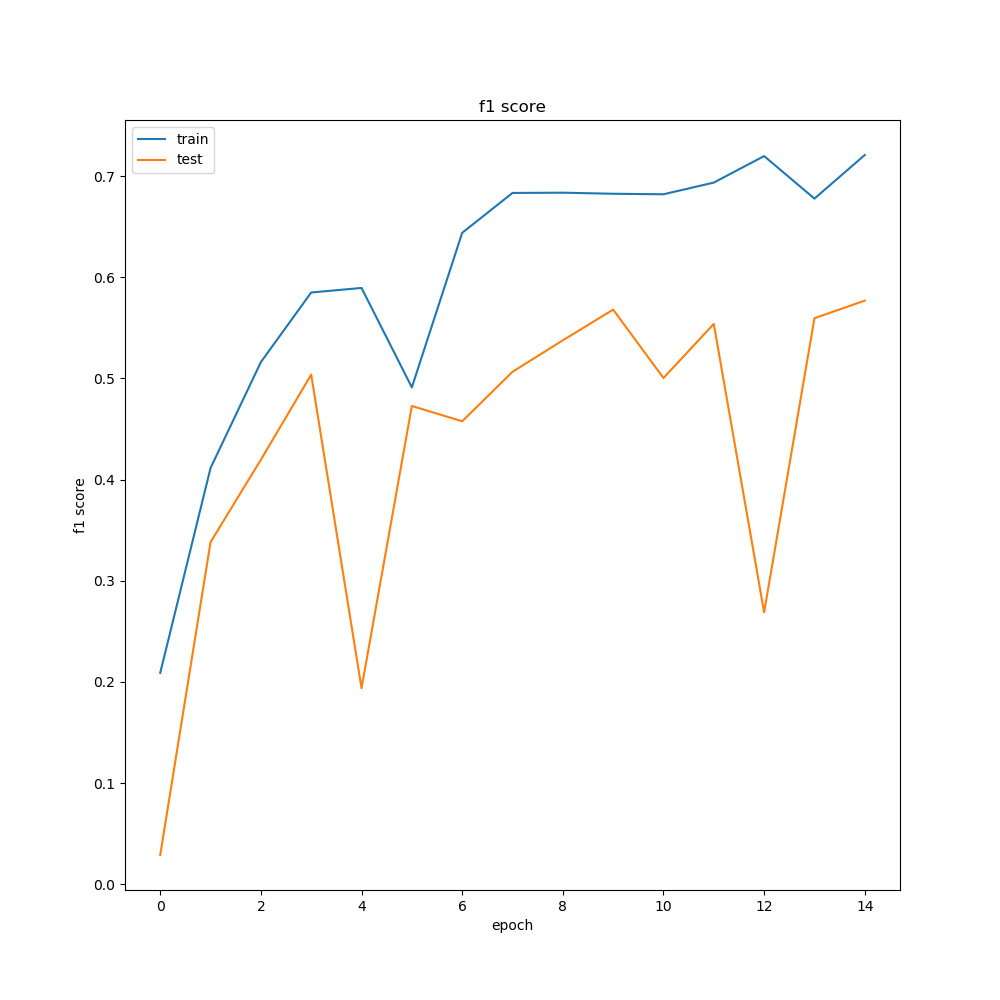}} &
\subcaptionbox{UNeXt3D}{\includegraphics[scale=0.23]{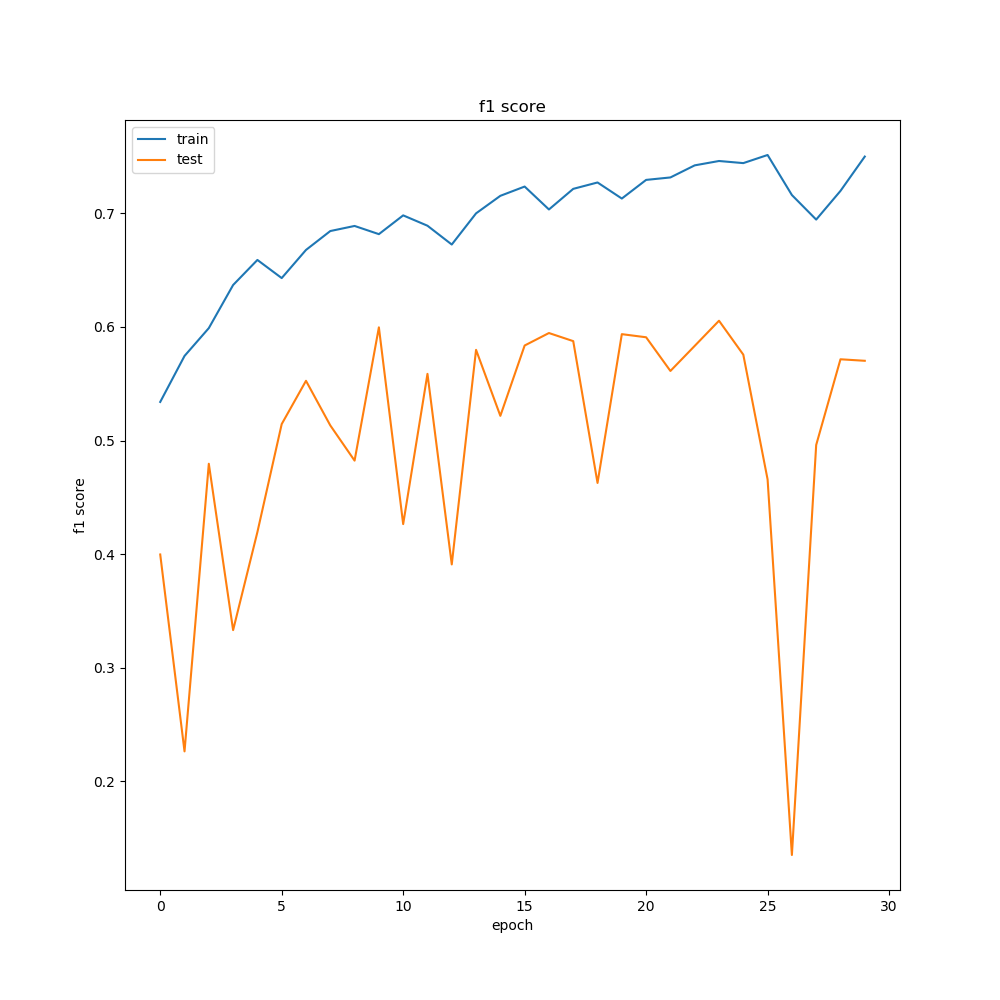}} &
\subcaptionbox{UNeXt3D with Inception}{\includegraphics[scale=0.23]{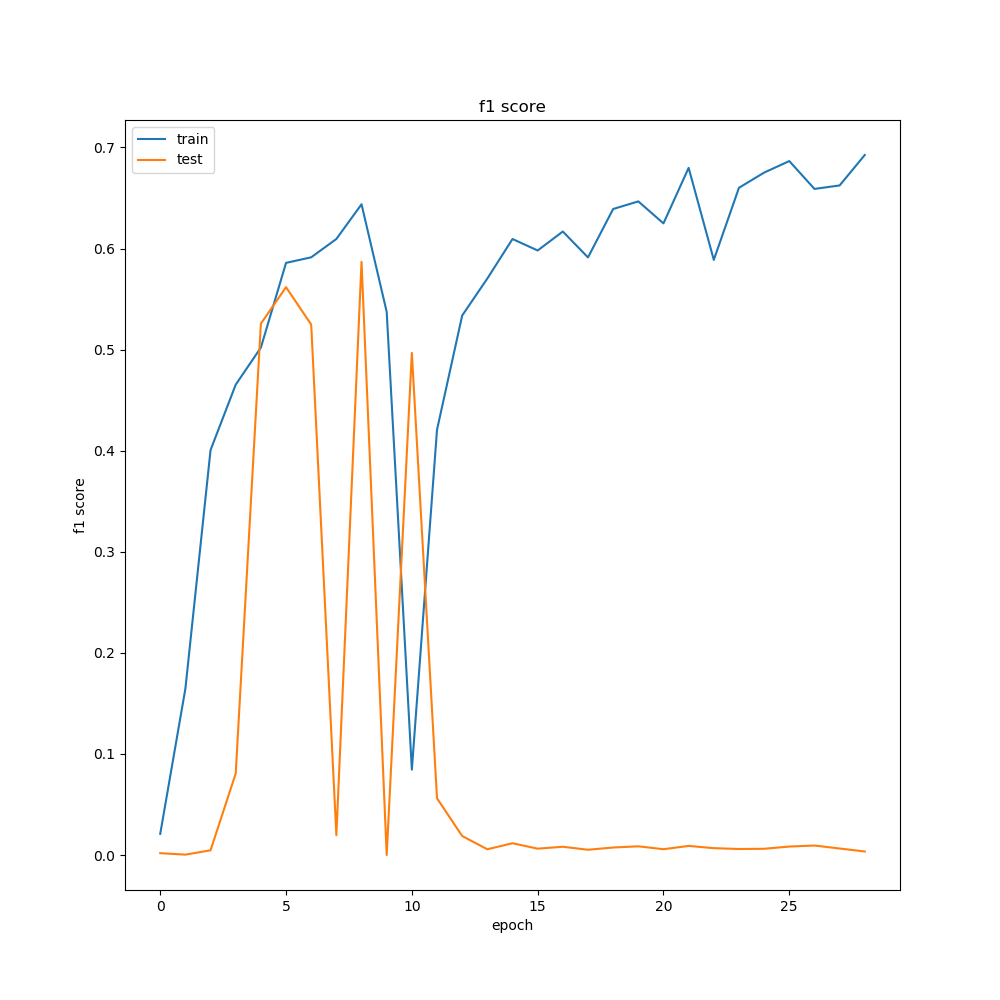}}
\end{tabular}
\caption{F1 Score Graphs for Train and Val for Each Model}
\label{fig:graphs}
\end{figure*}

\begin{figure*} 
\begin{center}
\resizebox{18cm}{!}{\begin{tabular}{ |c|c|c|c|c|c| }
 \hline
Model & Time Per Epoch & Num Params & Validation Accuracy & Train F1 Score & Validation F1 Score \\ 
 \hline
Baseline & 56.0148 & 70,853 & 0.9977 & 0.4442 & 0.433 \\
UNet3D & 287.2559 & 19,080,325 & 0.9982 & 0.7529 & 0.6189 \\
UNet3D Inception & 191.2815 & 2,033,713 & 0.9979 & 0.6731 & 0.6204 \\
SE-UNet3D & 296.0747 & 19,158,841 & 0.9977 & 0.7176 & 0.5903 \\
SE-UResNet3D & 352.4954 & 18,320,341 & 0.9981 & 0.7615 & 0.6348 \\
SE-UNet3D Inception & 186.015 & 637,404 & 0.9973& 0.5654 & 0.5273 \\
UResNet3D & 350.1806 & 26,453,317 & 0.9981 & 0.779 & 0.6367 \\
UNeXt3D & 337.1835 & 8,343,301 & 0.9981 & 0.7461 & 0.6055 \\
UNeXt3D Inception & 253.9381 & 2,434,909 & 0.998 & 0.6438 & 0.5869 \\
 \hline
\end{tabular}}
\end{center}
\end{figure*}

\begin{figure*}
	\centering
	\includegraphics[scale=1.2]{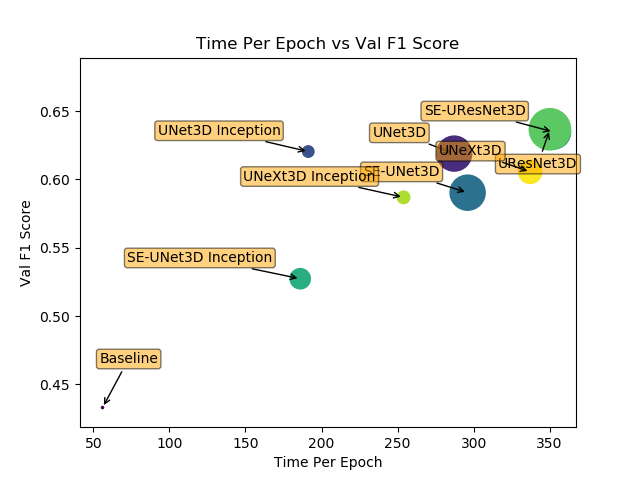}
    \caption{Circle size is based on number of parameters.}
  \label{fig:timevsf1}
\end{figure*}
\clearpage
}

%-------------------------------------------------------------------------
\subsection{UResNet3D} \label{ures_exp}

UResNet was our best performing model. This is likely because of the residual connections, which allowed for gradients to flow backwards more easily. However, the trade-off is that the model took longer on each epoch and is larger in size.

%-------------------------------------------------------------------------
\subsection{SE-UNet3D and SE-UNet3D with Inception} \label{se_unet_exp}

Adding the squeeze-and-excitation layers to UNet3D and UNet3D-Inception did not perform as well as the standard UNet3D model. We believe that this is likely due to the fact that the additional squeeze-and-excitation layers require more training and regularization than the standard model. As we can see in the graphs from figure \ref{fig:graphs}, the squeeze-and-excitation layers also seem to be causing the model to overfit to the training data. Thus, in our future work we intend to run the models for an additional 30 epochs with more regularization and other variance-reducing techniques. 

%-------------------------------------------------------------------------
\subsection{SE-UResNet3D} \label{se_uresnet_exp}

As the propagation of the gradient is made easier by residual layers, we thought it would be interesting to put residual layers and squeeze-and-excitation layers together. Again, this model did not appear to do nearly as well as UResNet without squeeze-and-excitation layers, but we believe with additional training time and regularization techniques it could improve.  

%-------------------------------------------------------------------------
\subsection{UNeXt3D and UNeXt3D with Inception} \label{unext_exp}

In order to run these models, we were limited by the model size, as using the Next aggregated transformation layers roughly doubled the number of parameters per layer. This forced us to halve the number of filters in order to run the model, which may have made the model too simple. We can see from the graphs in figure \ref{fig:graphs} that both runs had worse performance on the validation set than the original UNet3D model. The models also appear to be overfitting the dataset more heavily, with the training f1 score significantly higher than the validation score. This was likely due to both convolutions in the aggregated transformation layers learning very similar representations. 

%%%%%%%%% Conclustion
\section{Conclusion} \label{conclusion}

We conclude that recent advances in architectures for tasks such as image detection on the ImageNet dataset can be successfully applied to significantly different CNN architectures, such as the UNet3D, for significantly different tasks, such as Brain Tumor segmentation. This can be seen by the success of our UNet3D with inception layers and the Residual UNet3D models. We would like to point out that these models out performed the Vanilla UNet3D with very limited hyperparameter tuning. We choose to limit the amount of hyperparameter tuning to keep the models from diverging too much from each other. So that the meta architectural structures of the models could be directly evaluated against each other. We believe that further tuning of these models could lead to significant performance improvements, ultimately supplying a new set of tools for researchers in the biomedical field. 

%%%%%%%%% FUTURE WORK
\section{Future Work} \label{future}

Our primary focus for the future is working with the new 2018 dataset and submitting a model and a paper for the 2018 BraTS competition. In order to do so, we will alter our models to classify all five of the classes for the BRATS dataset and evaluate how our models perform in a multi-class classification problem. Additionally we will be further tuning each model architecture to fully understand its capabilities. Once we run our models on all of the classes, we also plan on extracting the features and predictions for each model to help us better visualize our comparison between models, as well as potentially learn new ways in which to improve upon these models.

%%%%%%%%% References
{\small
\bibliographystyle{ieee}
\bibliography{bib}
}

\end{document}